\documentclass{article}

\usepackage[sglblindworkshop, nonatbib, final]{neurips_2025}

\usepackage[utf8]{inputenc} 
\usepackage[T1]{fontenc}    
\usepackage{hyperref}       
\usepackage{url}            
\usepackage{booktabs}       
\usepackage{amsfonts}       
\usepackage{nicefrac}       
\usepackage{microtype}      
\usepackage{xcolor}         

\usepackage{graphicx}
\usepackage{multirow}
\usepackage{makecell}
\usepackage{amsmath}
\usepackage{subcaption}
\usepackage{pgfplots}
\usepackage[numbers]{natbib}
\usepackage[table]{xcolor}

\usepackage{enumitem}

\pgfplotsset{compat=1.18}

\definecolor{darkgreen}{RGB}{0,120,0}

\newcommand{\squishlist}{
 \begin{list}{$\bullet$}
  { \setlength{\itemsep}{0pt}
     \setlength{\parsep}{3pt}
     \setlength{\topsep}{3pt}
     \setlength{\partopsep}{0pt}
     \setlength{\leftmargin}{1.5em}
     \setlength{\labelwidth}{1em}
     \setlength{\labelsep}{0.5em} } }

\newcommand{\squishlisttwo}{
 \begin{list}{$\bullet$}
  { \setlength{\itemsep}{0pt}
     \setlength{\parsep}{0pt}
    \setlength{\topsep}{0pt}
    \setlength{\partopsep}{0pt}
    \setlength{\leftmargin}{2em}
    \setlength{\labelwidth}{1.5em}
    \setlength{\labelsep}{0.5em} } }

\newcommand{\squishend}{
  \end{list}  }

\workshoptitle{ML For Systems}
\title{APCE: Adaptive Progressive Context Expansion for Long Context Processing}

\author{%
 \textbf{Baisub Lee},
 \textbf{Sanghyun Byun},
 \textbf{Mohanad Odema},
 \\
 \textbf{Jung Guack},
 \textbf{Jacob Song},
 \textbf{Woo Seong Chung}
\\
\\
LG Electronics USA
}

\pgfplotsset{compat=1.18}

\begin{document}

\maketitle

\begin{abstract}


Deploying useful Long-Context Transformer Models (LCTMs) requires addressing two key challenges: (1) A \textit{growing memory footprint} due to quadratic self-attention and linear KV-cache scaling in memory as sequence length increases; (2) the \textit{ContextRot} phenomena where empirical evidence suggests that transformer architecture's performance degrades with increasing context length. Given the shared dependency on the input, a natural question arises: \textit{"Can we surgically select the most important input chunks for processing to synergistically (a) reduce the memory footprint, and (b) mitigate the ContextRot effects?"}
In this paper, we answer this question in the affirmative for \textit{long-context summarization} tasks. We propose APCE as a context-aware solution to select the most important input chunks through low-dimensional semantic similarity matching with the current query. By directly operating on the input, APCE decouples from strict dependency on underlying hardware or CUDA environments, promising a compatible solution scalable to different deployment systems. Our empirical evaluations have demonstrated superior or on-par summarization performance for APCE compared to the full dense baseline using a fraction (50\%-70\%) of the input sequence resulting in KV-cache and self-attention memory efficiency improvements. We hope our findings inspire further research on context-aware efficiency solutions for LCTMs geared towards other relevant long-context tasks. 






\end{abstract}
\section{Introduction}

Long-Context Transformer Models (LCTMs) are becoming increasingly popular given their ability to work with long sequences (2k - 1M context length) to enable long context tasks such as  long-document summarization, reasoning, and multi-modal understanding \cite{zaheer2020big, beltagy2020longformer, kryscinski2021booksum, alayrac2022flamingo, lin2023video}. In order to deploy useful LCTMs, two things to consider: (1) the \textit{growing memory footprint} of the model stemming from the quadratic memory complexity of self-attention operation, as well as the linear memory requirement for storing the KV-cache; (2) Addressing the \textit{ContextRot} \cite{hong2025context} phenomena, where empirical evidence has suggested that the transformer architecture struggles to maintain performance as context length increases -- especially challenging with current the scales (2k - 1M context length).

Sparsification and KV cache compression methods have been heavily studied to address these challenges (See Appendix \ref{sec:related_works}). However, two aspects remain relatively underexplored: (1) Most sparsification methods focus their innovations on the attention operation, leveraging low-dimensional importance calculations for the \textit{Key} and \textit{Value} tokens/blocks to drive the sparsification decisions. Though effective, the caveat remains that the \textit{Key} and \textit{Value} blocks from every input token/block in every self-attention operation would require loading into the lower memory levels at least once for the low-dimensional projection and importance computations, necessitating specialized kernel support to compensate for the added operations and data movement overheads; (2) Eviction strategies for KV cache compression are \textit{persistent}, lacking the means to reconsider evicted tokens/blocks even if they become more relevant as the context evolves \cite{tang2024quest, xu2025xattention}. This constitutes a common situation for long-context scenarios (e.g., spread out dependencies across multiple passages in long-documents).

We present APCE (Adaptive Progressive Context Expansion) as a complementary method to compensate for these two points by directing sparsification efforts towards the input chunks themselves while allowing Reprioritization to update chunks selection. Through a low-dimensional semantic similarity matching approach, APCE can determine which input chunks are most important in relation to the current query, choosing only those to proceed with the downstream transformer operations. Since APCE operates on the input, the low-dimensional representations of the input chunks are only computed once during prefill, and kept in memory for quick similarity matching operations enabling chunk selection update as the context evolves. We summarize the paper's contributions as follows:


\begin{itemize}
    \item We propose APCE, a context-aware input chunk sparsification solution leveraging semantic similarity matching and supporting chunk replacement for efficient long-context processing.
    \item We evaluate APCE on long-context summarization task, demonstrating similar or superior performance to that of the full dense baseline using a fraction (50-70\%) of input chunks.
    \item We perform ablation studies on APCE's Reprioritization to characterize its performance-efficiency trade-offs, and provide analytical analysis on APCE's memory efficiency gains. 
    \item We discuss future research potential for scaling APCE to further related long-context tasks.

\end{itemize}

\section{Method} \label{sec:method}

\begin{figure}[!t]
\begin{center}
{\includegraphics[,width = 0.95\textwidth]{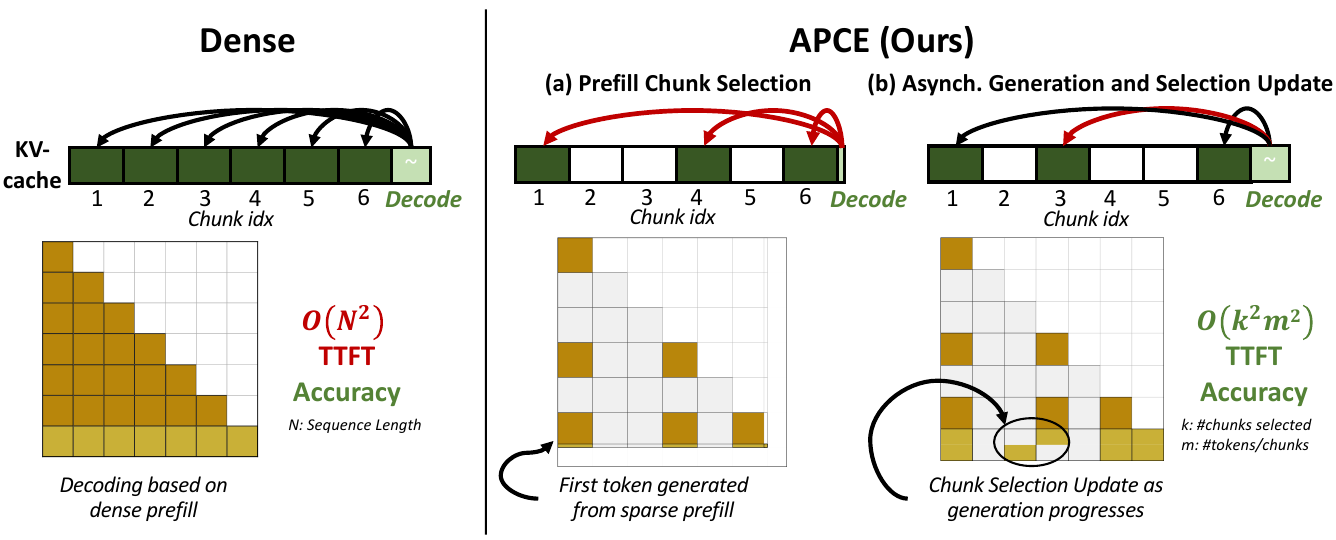}}
\end{center}
\vspace{-2ex}
\caption{Overview of APCE for an example with 6 chunks. (\textit{left}) The full dense baseline attention suffers from quadratic complexity and long time-to-first-token (TTFT). (\textit{middle}) APCE performs prefill chunk selection using semantic similarity, resulting in sparse attention reducing complexity and TTFT. (\textit{right}) Chunk selection is updated as generation progresses to maintain performance.}
\label{fig:APCE}
\end{figure}

\textbf{Base Formulation. }Denote $s=[x_1, x_2, \cdots, x_N]$ as a sequence of $N$ input tokens to be processed where $x_i \in \mathbb{N}$ represents the $i^{th}$ token. For long context, this token sequence can be partitioned into chunks such that the $i^{th}$ chunk is defined as $x^{(c_i)} = [x_1^{(c_i)}, x_2^{(c_i)}, \cdots, x_{m_i}^{(c_i)}]$, with entry $x_j^{(c_i)} \in \mathbb{N}$ represents the $j^{th}$ token belonging to a chunk $i$ of size $m_i$. 

Let $f(\cdot)$ be a projection function for mapping token sequences into low sequence embeddings where $\mathbf{c_i} = f(x^{(c_i)})$, $\mathbf{c_i} \in \mathbb{R}^d$ represents a $d$-dimensional embedding representation of the $i^{th}$ chunk.
Given a sequence of $n$ chunks, denote $\mathbf{C} = [\mathbf{c_1}, \mathbf{c_2}, \cdots, \mathbf{c_n}]$ as their corresponding sequence of $d$-dimensional embedding representations. Define also the query token sequence containing an instruction or question as $x^{(q)} = [x_1^{(q)}, x_2^{(q)}, \cdots, x_{k}^{(q)}]$. From it, we also obtain $\mathbf{q} = f(x^{(q)}), \mathbf{q} \in \mathbb{R}^d$ as the $d$-dimensional query embedding.
After chunked prefill, APCE targets identifying the indices for the $top$-k most relevant chunks maximizing a similarity function:
\[
\mathcal{I}_{\text{top-}k}
= \underset{i \in \{1, \dots, n\}}{\operatorname{arg\,top}\,k} \;
\mathrm{Sim}(\mathbf{q}, \mathbf{c}_i)
\]
where $\mathrm{Sim}(\mathbf{q}, \mathbf{c}_i)$ represent the similarity score between the current query and the $i^{th}$ chunk. The top-k selected chunks progress for processing and self-attention computation as shown in Figure \ref{fig:APCE}. Throughout generation, $\mathbf{C}$ embeddings remain cached, while $\mathbf{q}$ is continously cached and updated.

\textbf{Similarity Function. } To capture semantic similarity between embeddings, \textit{Semantic Scoring} via cosine similarity can be adopted similar to BERTScore \cite{bertscore2019}:
    \[
\mathrm{Sim}(\mathbf{q}, \mathbf{c}_i) 
= \frac{\mathbf{q}^\top \mathbf{c}_i}{\|\mathbf{q}\|_2 \, \|\mathbf{c}_i\|_2}
\]    

\textbf{Reprioritization. }To maintain relevancy of selected chunks with regards to the current query, APCE supports reprioritization to re-evaluate chunk similarity scores with current embedding and update the chunk selection accordingly. Reprioritization operates with the following two features:

\begin{itemize}
    \item \textit{Buffer Management}. Based on the max chunk selection size, lowest scoring chunk representations in the buffer can be evicted to make way for the new, higher scoring chunks.
    \item \textit{Recomputation}. We support recomputing the K and V values for any chunks in the buffer if their attention scores depend on a newly selected chunk(s). This presents a trade-off between mitigating attention inconsistencies due to out-of-order chunk loading, and the added computational overhead. The ablation study in Appendix \ref{appdx:reprioritization} touches on this trade-off.
\end{itemize}

\textbf{Enhanced Query Embedding. }As decoding progresses, APCE dynamically updates the query representation by blending in context from both the original instruction and recently generated tokens.
The enhanced query is updated every Reprioritization Interval.

\textbf{Asynchronous Generation. }We extend APCE to support asynchronous token generation to improve Time-to-First-Token (TTFT) in cases where constrained system resources (I/O, memory bandwidth) enforce a slow prefilling. APCE can initiate token generation from partially loaded chunks, progressively improving output generation quality as more context is acquired with more loaded chunks.










\textbf{Complexity analysis. } For a sequence length of $N$ tokens, $k$ selected chunks, and $m$-token chunk size, APCE reduces attention complexity from $\mathcal{O}(N^2)$ to $\mathcal{O}(k^2m^2)$ especially viable when $km \ll N$.

\section{Experiments}

\begin{table}[!t]
\centering
\footnotesize
\caption{Aggregate Statistics for BookSum \cite{kryscinski2021booksum}  and timings comparisons for APCE on Llama-3.2-3B-Instruct \cite{dubey2024llama} across all context lengths when varying the \textbf{maximum number of chunks selected}. Higlighted cells are ones where APCE achieves the best numbers.}
\vspace{2ex}
\label{tab:boosum_summary}
\begin{tabular}{ccccc}
\toprule
\multirow{2}{*}{Method} 
  & \multicolumn{2}{c}{Performance $\uparrow$} & \multicolumn{2}{c}{Timings $\downarrow$} \\
\cmidrule(lr){2-3} \cmidrule(lr){4-5}
& BERTScore (F1) & ROUGE-L (F1) & TTFT (s) & Total time (s) \\
\midrule
  Dense Baseline & 0.8413±0.0005 & 0.1591±0.0082 & 4.6820±2.9131 & 14.8356±6.7588 \\
  \midrule
  \textbf{APCE (50\% chunks used)}    & 0.8412±0.0014 & 0.1521±0.0028 & \cellcolor{green!20}2.8161±1.7641 & 16.4040±7.6996 \\
\midrule
  \textbf{APCE (60\% chunks used)}      & 0.8400±0.0027 & 0.1567±0.0096 & 3.3041±1.9478s & 17.9446±7.9049 \\
\midrule
  \textbf{APCE (70\% chunks used)} & \cellcolor{green!20}0.8418±0.0010 & 0.1584±0.0078 & 3.6693±2.1177 & 19.6179±8.6174 \\
\bottomrule
\end{tabular}
\end{table}

We evaluate the performance of APCE on long-context summarization task. We provide a summarized version of the results here and refer the reader to Appendix \ref{appdx:results} for full experiments and setup details. 

\textbf{Experimental Details. }Our evaluations are conducted using \textit{BookSum} \cite{kryscinski2021booksum}, a large collection of datasets for long-form narrative summarization. We specify 3 context length groups (8k, 20k, and 30k), and select representative test samples from each groups to conduct our APCE evaluations at different context lengths. We use a Llama-3.2-3B-Instruct \cite{dubey2024llama} as our base model, comparing APCE's performance against the baseline Full Dense model using an NVIDIA RTX 4090 with 24 GB memory.



\begin{figure}[!t]
\begin{center}
{\includegraphics[,width = 0.98\textwidth]{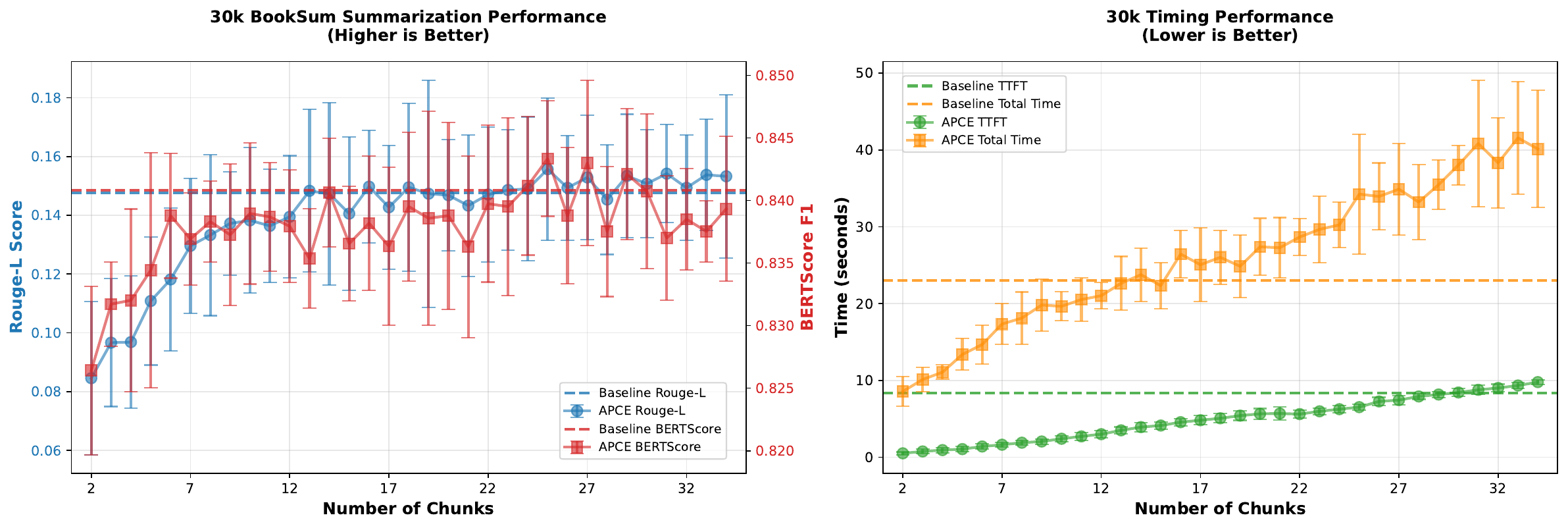}}
\end{center}
\vspace{-2ex}
\caption{Summarization Performance scores (\textit{left}) and inference timings (\textit{right}) for APCE compared to the baseline when varying the number of loaded chunks on Llama-3.2-3B-Instruct \cite{dubey2024llama} the 30K context length group from BookSum \cite{kryscinski2021booksum}. Chunk size is fixed to 800 tokens.}
\label{fig:30k_chunks}
\end{figure}

\textbf{Experiments Scope. }
Our experimental analysis for APCE is performed by evaluating summarization performance, latency timings, and memory efficiency. We vary two hyperparameters in our experiments: (1) Maximum \#chunks selected given a fixed chunk size; (2) Chunk size given a fixed maximum \#chunks selected. We also perform ablations on the repriortization interval.


\subsection{Results Summary }

\textbf{Number of Selected Chunks Experiment. } Table \ref{tab:boosum_summary} shows aggregate summarization results across all context length groups when varying the number of chunks selected given 800 token chunk size. The Table shows APCE scores when 50\%, 60\%, and 70\% of input chunks are selected at a time. Generally, we observe that APCE is capable of maintaining comparable scores with the dense baseline across the BERTScore \cite{bertscore2019} and ROUGE-L \cite{rouge2004} metrics. APCE with 70\% selection outperforms the baseline on BERTScore while being within $\sim$0.5\% score margin on ROUGE-L score. Figure \ref{fig:30k_chunks} (left) shows APCE starts to exhibit on-par, occasionally superior, performance to the baseline using $\sim50$\% of the input chunks for the 30k group. Figure \ref{fig:apce_master_nchks} shows the full results across all groups.


\textbf{Chunk Size Experiment. }Table \ref{tab:boosum_summary_cksize} shows the aggregated results when varying the chunk size given 70\% of available chunks are selected. The results show 800 token chunk size can lead to on-par performance with the dense baseline, further corroborating the importance of choosing the right input chunks. Figure \ref{fig:apce_master_chksize} illustrates the full results breakdown across the different context lengths.

\textbf{Reprioritization and Memory Analysis. }Appendix \ref{appdx:reprioritization} introduces Reprioritization ablation study showing the performance-efficiency trade-off affecting total generation time. Appendix \ref{appdx:memory} provides an analytical analysis on APCE's memory efficiency gains -- reaching 55.6\% for prefill attention.

\section{Conclusion and Future Directions}

We presented APCE, an input chunk sparsification solution for LCTMs improving their performance and memory efficiency on long-context summarization tasks. We highlight the following takeaways: 

\textbf{Beyond Summarization.} APCE has demonstrated promising results on long-context summarization, paving the way to investigate its applicability in other long-context task scenarios (long-context reasoning and multi-turn dialogues). Considering their task-specific challenges, we hypothesize another level of sophistication is needed to preserve semantic understanding as input is sparsified.

\textbf{APCE is a complementary Solution. } Given its focus on the input chunks directly, APCE is compatible with existing SOTA solutions that adopt sparsification in the self-attention operation, enabling delivering performance and efficiency improvements from two dimensions.

\textbf{On the implementation side.} Our vanilla APCE solution demonstrated promising results. More sophisticated implementations can consider context-boundary awareness for semantic similarity matching and supporting compute kernels for efficiency. 

\textbf{Navigating Performance-Efficiency Trade-offs. } Reprioritization Interval can be adjusted to meet the target use-case requirements with regards to performance and total generation time. Whereas the reduced TTFT can benefit use cases where response time is critical (e.g., low-risk robotic interactions).



\bibliographystyle{plainnat}
\bibliography{ref}

\appendix

\section{Related Works}\label{sec:related_works}

Long Context models are becoming increasingly popular to accommodate emerging multi-modal tasks \cite{alayrac2022flamingo}, long context reasoning, and summarization \cite{longbench2023}. Recent SOTA LLM models like Llama \cite{dubey2024llama} support Rotary Position Embeddings (RoPE) \cite{su2024roformer} to scale their context window up to 128K or 1M tokens \cite{google2025gemini25}. As the sequence length increases, the complexity of attention operation also scales quadratically, posing efficiency challenges for LLM inference. Thus, recent works propose sparse attention techniques to address these challenges \cite{beltagy2020longformer, zaheer2020big, xu2025xattention, tang2024quest, wang2021spatten, jiang2024minference, lai2025flexprefill, an2024training, zhang2023h2o, xiao2023streamingllm, oren2024transformers, wang2025llms}.

Initially, sparsification works adopted static predefined sparse attention patterns \cite{beltagy2020longformer, zaheer2020big} that reduce the attention computation requirements. However, the inflexibility of static patterns has led recent works to shift towards dynamic compute sparsity where irrelevant attention computations are skipped at runtime through masking/blocking techniques \cite{xu2025xattention, tang2024quest, wang2021spatten, jiang2024minference, lai2025flexprefill, an2024training}. For example, X-Attention \cite{xu2025xattention} use anti-diagonal scoring to identify irrelevant attention blocks, and prune them from the attention structure to improve the computational footprint, whereas Quest \cite{tang2024quest} relies on similarity matching between current query vector and vector-wise statistics of each KV page. Notably, the implementation of these methods has been CUDA-centric, where their efficiency is dependent on hardware-specific custom kernel (e.g., FlashAttention \cite{dao2022flashattention}) that can implement the masking/blocking for sparse attention. 




Another body of works targeted KV cache management strategies \cite{zhang2023h2o, xiao2023streamingllm, oren2024transformers, wang2025llms} where low-scoring tokens are evicted from the cache to improve attention efficiency. For instance, H20 \cite{zhang2023h2o} retains a mixture of recent tokens and those with the highest cumulative attention scores (heavy hitters); whereas StreamingLLMs \cite{xiao2023streamingllm} leverages sink tokens and a sliding window to maintain most relevant tokens. These methods still rely on loading the full KV-cache initially in memory for scoring prior to the application of any eviction, and once a token is discarded it cannot be recovered even when its relevance increases as generation evolves (as was remarked in \cite{tang2024quest}).

The differences in APCE's approach are twofold: (1) it adopts sparsification in the input stage, enabling it to accelerate both the prefilling and decoding through semantic similarity without the need for computing initial attention scores or statistics for token eviction decisions; (2) It supports chunks replacement to maintain performance as the long context continues to evolve.

\section{Detailed Experiments} \label{appdx:results}

\subsection{Experimental Setup }

\textbf{Long-Context Summarization Dataset. }Our long-context summarization experiments are conducted on \textit{BookSum} \cite{kryscinski2021booksum}. \textit{BookSum} is a large collection of datasets for long-form narrative summarization containing a variety of long-context inputs (e.g., books, chapters, paragraphs). To cover different ranges of context lengths, we first define three input lengths groups with base lengths of 8k, 20k, and 30k. Then, we assign to each group the test samples from \textit{BookSum} that are within 10\% margin of the corresponding base input length. From each group, we select the first 10 input samples for our final evaluations. We remark that in the 30k group, there were only 6 test samples in the group and thus were all considered in the evaluations. 

\textbf{Evaluation Metrics} We use the following metrics for our summarization evaluations:
\begin{itemize}
    \item \textbf{ROUGE-L}~\citep{rouge2004}: Evaluates longest common subsequence overlap, emphasizing sentence-level structural similarity.
    \item \textbf{BERTScore}~\citep{bertscore2019}: Measures semantic similarity using contextual embeddings, providing more nuanced evaluation of meaning preservation.
\end{itemize}

\textbf{Baselines, Model, and Hardware. }We evaluate APCE and compare against a Full-dense baseline that uses all the input token sequence for its attention computation and downstream processing. We implement both solutions on a Llama-3.2-3B-Instruct \cite{dubey2024llama}. We perform our experiments by default on an NVIDIA RTX 4090 with 24 GB memory. 

\textbf{Experiments Description. }
Our main evaluations on APCE involve varying two hyperparameters: (1) The \#chunks selected given a fixed chunk size; (2) The chunk size given a fixed \#chunks selected. When varying the \#chunks selected, we fix the chunk size at 800 tokens. When varying the chunk size, we set the \#chunks selected to the equivalent number around 70\% of sequence length (7, 18, and 24 for respective 8k, 20k, and 30k context-length groups). We also perform an ablation study on the effects of Reprioritization Interval, and conduct an analytical memory efficiency analysis for APCE compared to the Full Dense.  

\textbf{APCE Configuration. }
By default, we set for APCE the Reprioritization Interval to 50 tokens with Recomputation enabled. We set asynchronous generation to start after loading the 4th chunk. For the enhanced query embedding during generation, we project the last 100 characters in the question/instruction, and blend them with the last 50 generation tokens.

\textbf{Chunks' Embedding Projection. }
We project each chunk into a low-dimensional embedding of size 384. We use SentenceTransformer \cite{all-MiniLM-L6-v2} for our projection. The memory overhead is minimal compared to other components since the projections are stored only once to used across all self-attention layers. For instance, projections of 37 chunks of size 800 tokens (equivalent to the 30k context length) result in $\sim$28 MB memory footprint in FP16 format.

\subsection{BookSum Summarization Experiments} \label{appdx:booksum}

\begin{table}[!t]
\centering
\footnotesize
\caption{Aggregate Statistics for BookSum \cite{kryscinski2021booksum} performance and timings comparisons for APCE on Llama-3.2-3B-Instruct \cite{dubey2024llama} across all context lengths when varying the \textbf{Chunk Size}. Higlighted cells are ones where APCE achieves the best numbers.}
\vspace{2ex}
\label{tab:boosum_summary_cksize}
\begin{tabular}{ccccc}
\toprule
\multirow{2}{*}{Method} 
  & \multicolumn{2}{c}{Performance $\uparrow$} & \multicolumn{2}{c}{Timings $\downarrow$} \\
\cmidrule(lr){2-3} \cmidrule(lr){4-5}
& BERTScore (F1) & ROUGE-L (F1) & TTFT (s) & Total time (s) \\
\midrule
  Dense Baseline & 0.8395±0.0016 & 0.1610±0.0032 & 5.1497±3.1950 & 16.0875±8.0018 \\
  \midrule
  \textbf{APCE (800 token chksize)}    & \cellcolor{green!20}0.8419±0.0026 & 0.1606±0.0098 & \cellcolor{green!20}3.9516±2.5660 & 19.0695±10.2352 \\
\midrule
  \textbf{APCE (1000 token chksize)}      & 0.8398±0.0015 & 0.1548±0.0095 & 4.7861±2.9681 & 20.1979±9.5202 \\
\midrule
  \textbf{APCE (1600 token chksize)} & 0.8295±0.0158 & 0.1472±0.0219 & 6.3236±3.8602 & 27.3208±16.9383 \\
\bottomrule
\end{tabular}
\end{table}

\begin{figure}[p]
\begin{center}
{\includegraphics[,width = 0.98\textwidth]{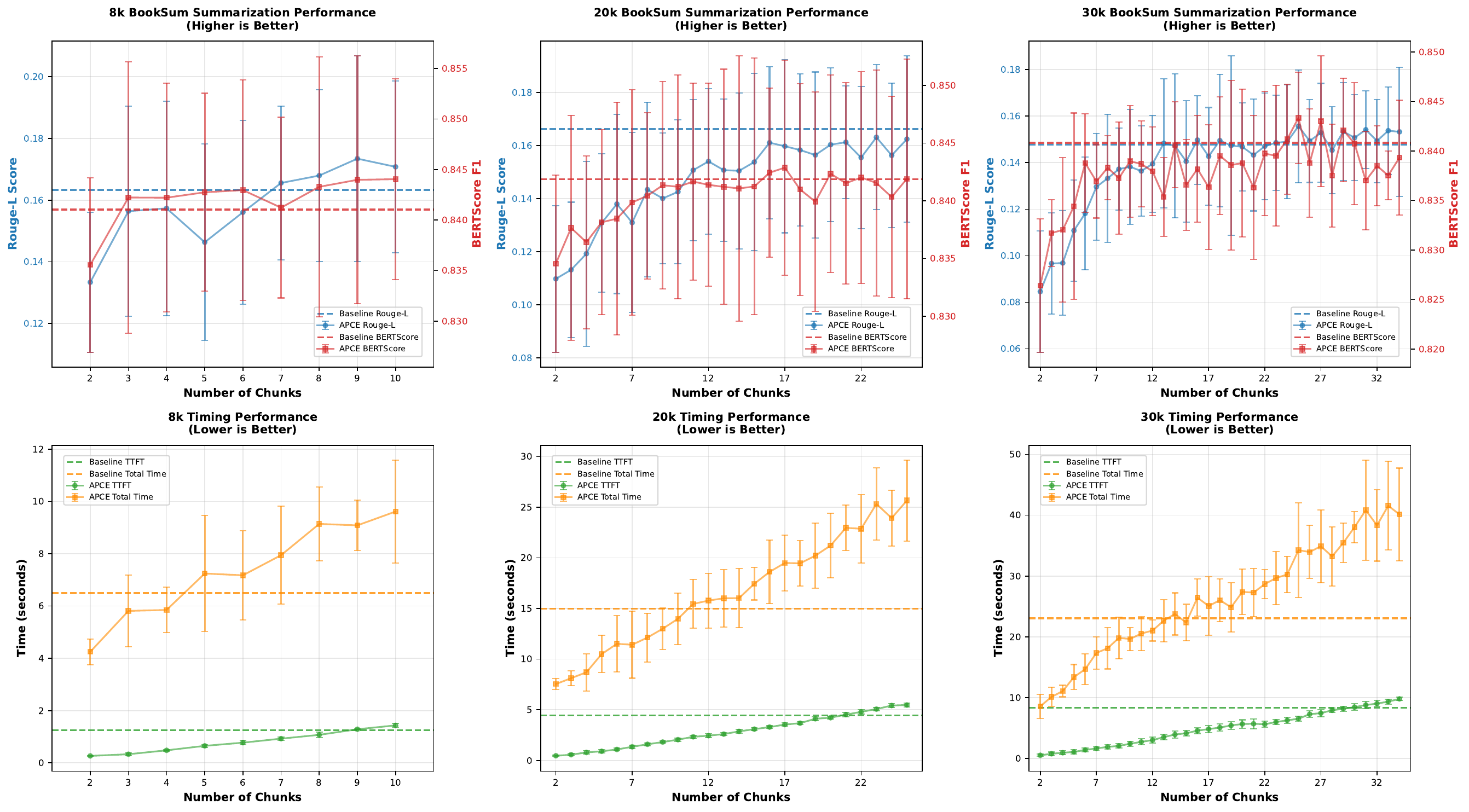}}
\end{center}
\vspace{-2ex}
\caption{Comparing APCE and the Full Dense baseline across the 3 groups of long context when \textbf{varying the number of chunks at fixed chunk size of 800}. The top row shows the performance results on BookSum \cite{kryscinski2021booksum}, whereas the bottom row shows the timing evaluations. }
\label{fig:apce_master_nchks}
\end{figure}
\begin{figure}[p]
\begin{center}
{\includegraphics[,width = 0.98\textwidth]{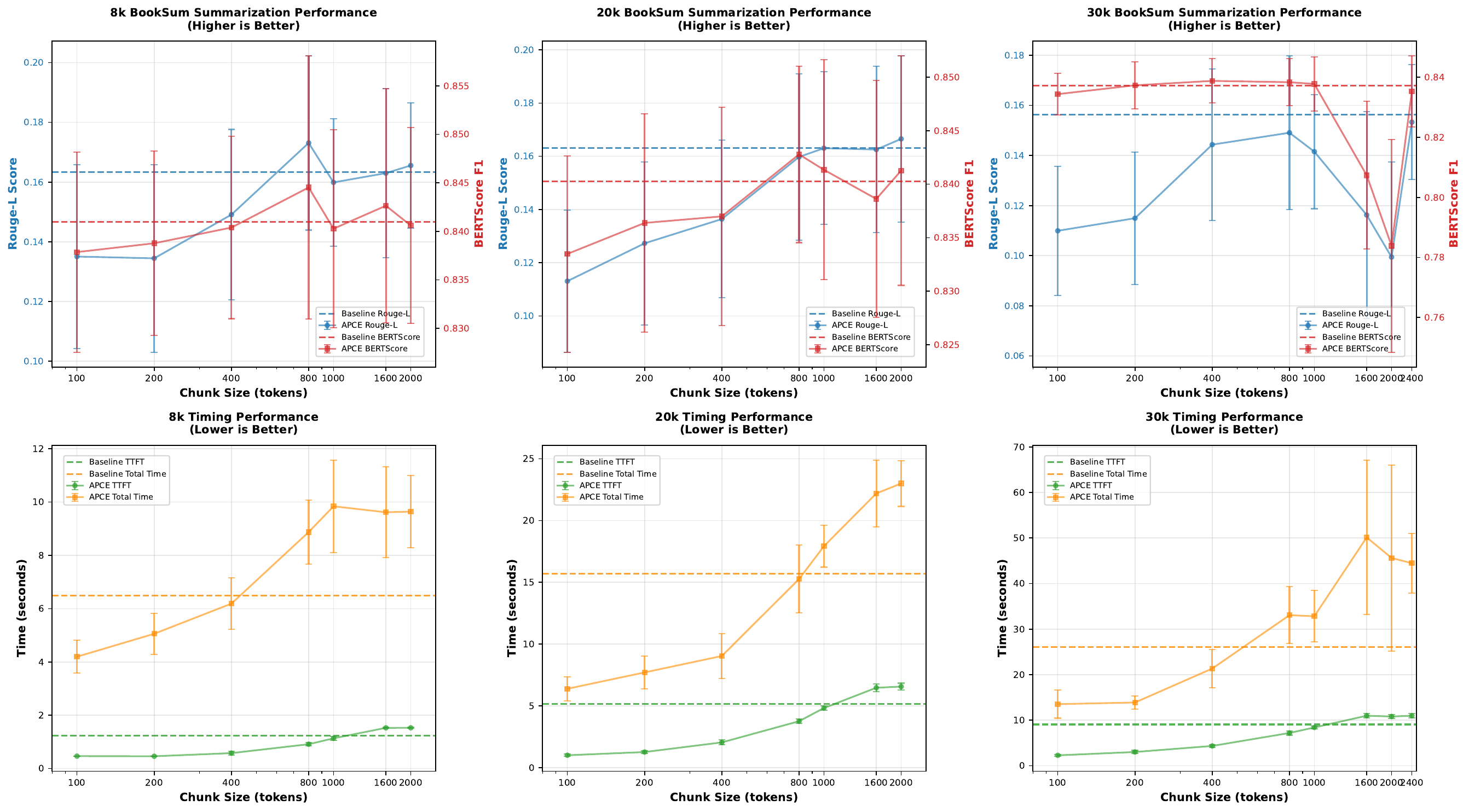}}
\end{center}
\vspace{-2ex}
\caption{Comparing APCE and the Full Dense baseline across the 3 groups of long context when \textbf{varying the chunk size at 70\% of available chunks selected}. The top row shows the performance results on BookSum \cite{kryscinski2021booksum}, whereas the bottom row shows the timing evaluations. }
\label{fig:apce_master_chksize}
\end{figure}

We evaluate the long-context summarization performance of APCE and the Full Dense baseline on BookSum \cite{kryscinski2021booksum} across the 3 context length groups (8k, 20k, 30k). Tables \ref{tab:boosum_summary} and \ref{tab:boosum_summary_cksize} provide an aggregated results in terms of summarization performance and timing evaluations across all context lengths covering the two respective experimental settings: (a) varying the number of chunks given a fixed chunk size (800 tokens); (b) varying the chunk size given a fixed number of chunks (70\% of total context length). The full results are provided in Figures \ref{fig:apce_master_nchks} and \ref{fig:apce_master_chksize}.

\textbf{Number of Chunks Variation. }As observed in Figure \ref{fig:apce_master_nchks}, APCE is capable of maintaining comparable scores with the dense baseline across the BERTScore \cite{bertscore2019} and ROUGE-L \cite{rouge2004} metrics across all context length groups starting from low number of chunks (around $\sim$30\% of the context length). At APCE with 70\% chunk usage, we find that APCE outperforms the baseline on BERTScore while being within $\sim$0.5\% score margin on ROUGE-L score (see Table \ref{tab:boosum_summary}). As for latency, our vanilla APCE implementation sustains longer generation times than the dense baseline when $\sim$50\% of chunks are selected. Key reasons include the added overhead of Reprioritization, which can be managed through (1) adjusting the Reprioritization Interval as shown in the ablation in Appendix \ref{appdx:reprioritization}. (2) Exploring more efficient APCE implementation with a supporting, customized self-attention operator (Future research point).
As for TTFT, asynchronous generation leads APCE to achieve speedups over the baseline, which can for its own part be useful for a special subset of applications (low-risk robotics). 

\textbf{Chunk Size Variation. }Figure \ref{fig:apce_master_chksize} shows that across the 3 context length groups, chunk size around 800-1000 token provide the best Rouge-L and BERTScore performance, higlighting the importance of semantic-based selection to improve performance. We also observe performance drop spikes (Figure \ref{fig:apce_master_chksize} top-right sub-figure at the 30k context length), which can generally attribute to one of two reasons: (1) \textit{ContextRot} as discussed in prior sections; (2) The chunking strategy resulted in a split that disrupted a coherent sequence, causing loss of knowledge. The latter point opens the door for further investigation on context-aware chunking strategies. Table \ref{tab:boosum_summary_cksize} provides an aggregated summary of the results across all context length groups for the 800, 1000, and 1600 chunk sizes. The timing plots exhibit similar patterns to the previous experiment and thus do not alter our conclusions.

\subsection{Ablation Study on Reprioritzation} \label{appdx:reprioritization}

We perform an ablation study on the Reprioritization interval using 10 books of 10k context length. To better observe the Reprioritization effects, we alter the configuration setting token chunk size to 256 tokens and the maximum number of chunks to 16. We show the results in Table \ref{tab:rep_256}, where the average total output generation length is 198.87±36.67 tokens. The results displayed in the table are the average across all test cases.

The first observation is that looking at the extreme replacement case (every 1 token), only 10.07\% of replacement opportunities are used.This suggests that chunk replacements around $\sim$10\% of the total output generation length (200-250 tokens in this experiment) are sufficient to maintain performance at levels comparable to the baseline. If we test this hypothesis and observe the reprioritization interval at 25 tokens, we observe that it is the only configuration that outperforms the dense baseline on both the Rouge-L \cite{rouge2004} and BERTScore \cite{bertscore2019} summarization metrics. 

As for performance-efficiency trade-offs, we observe that more frequent replacements (smaller reprioritization intervals) sustain more total generation time, which is to be expected in a vanilla APCE implementation considering the overhead of K and V recomputation.  Larger reprioritization intervals incur less overhead and get closer to the dense baseline total generation latency (6.64 seconds at 50 tokens). The tolerance of this trade-off depends on the underlying long-context use cases, and the potential for further APCE speedups remains open with a more sophisticated implementation.

\begin{table}[t]
\centering
\footnotesize
\caption{Ablation on Reprioritization Interval showing the average results for 10 book samples at 10k context length at \textbf{256 token chunk size}. Rep. Interval at 200 is closest to No Replacement case. Highlighted numbers are the ones discussed in the text.}
\label{tab:rep_256}
\vspace{2ex}
\resizebox{0.98\linewidth}{!}{
\begin{tabular}{c|cc|cc|cc}
\toprule
\textbf{Rep. Intvl} & \textbf{Rouge-L} & \textbf{BERTScore} & \textbf{TTFT} & \textbf{Total Time} & \textbf{Replace.} & \textbf{Replace.} \\
\textbf{(tokens)} & (F1) & (F1) & (s) & (s) & \textbf{taken} & \textbf{Available} \\
\midrule
1   & 0.1237±0.0282 & 0.8234±0.0160 & 0.861±0.02 & 11.551±4.97 & \cellcolor{green!20}20.1±8.0 & \cellcolor{green!20}198.5±123.6 \\
5   & 0.1435±0.0179 & 0.8182±0.0255 & 0.552±0.07 & 11.409±4.78 & 13.0±4.6 & 50.6±27.7 \\
10  & 0.1364±0.0357 & 0.8186±0.0233 & 0.542±0.07 & 9.309±3.76 & 9.2±2.2 & 20.4±12.2 \\
25  & \cellcolor{green!20}0.1360±0.0294 & \cellcolor{green!20}0.8331±0.0110 & 0.549±0.07  & 7.802±1.09 & 4.7±1.8 & 6.5±1.1 \\
50  & 0.1219±0.0249 & 0.8316±0.0094 & 0.549±0.07 & \cellcolor{green!20}6.640±0.51 & 1.9±0.7 & 2.4±0.5 \\
100 & 0.1355±0.0264 & 0.8200±0.0300 & 0.543±0.07 & 8.608±2.55 & 1.2±0.6 & 1.8±1.0 \\
200 & 0.1246±0.0314 & 0.8211±0.0333 & 0.543±0.07 & 7.431±2.03 & 0.2±0.4 & 0.4±0.7 \\ \midrule
Dense Baseline & 0.1270±0.0147 & 0.8240±0.0123 & 1.731±0.11 & 6.360±0.83 & - & -\\
\bottomrule
\end{tabular}}
\end{table}

\section{Memory Footprint Analysis }\label{appdx:memory}
We provide an analytical analysis of the theoretical memory efficiency gains from APCE.

\subsection{Notation}
\begin{itemize}
    \item $d_{\text{emb}}^{(Q)}$: Query and output embedding dimension
    \item $d_{\text{emb}}^{(KV)}$: Key and Value embedding dimension (aggregate across attention heads)
    \item \text{seq\_len}: sequence length
    \item \text{chunk\_size}: APCE chunk size
    \item \text{n}\_{\text{chunks}}: APCE number of chunks
\end{itemize}



\subsection{KV-cache }

For attention computation, the low-level KV cache requirement for dense and APCE is given by:
\[
\text{KV}_{\text{dense}} \;=\; \text{seq\_len} \times 2 \times d_{\text{emb}}^{(KV)}
\]
\[
 \text{KV}_{\text{APCE}} \;=\; \text{n}_{\text{chunks}} \times \text{chunk\_size} \times 2 \times d_{\text{emb}}^{(KV)}
\]

\subsection{Self-Attention }

The self-attention memory requirement entails reserving space for (a) Q, K, V vectors, (b) attention matrix (quadratic sequence length), and (c) the output d-dimensional token vector. For the prefill, the self-attention memory requirement is given as:
\[
\text{Mem}_{\text{dense\_prefill}} \;=\; 
\big( 2 \cdot \text{seq\_len} \cdot d_{\text{emb}}^{(KV)} + 2 \cdot \text{seq\_len} \cdot d_{\text{emb}}^{(Q)} \;+\; \text{seq\_len}^{2} \big)
\]
\[
\text{Mem}_{\text{APCE\_prefill}} \;=\; 
\Big( 2 \cdot (\text{n}_{\text{chunks}} \cdot \text{chk\_size}) \cdot d_{\text{emb}}^{(KV)} + 2 \cdot (\text{n}_{\text{chunks}} \cdot \text{chk\_size}) \cdot d_{\text{emb}}^{(Q)}
\;+\; (\text{n}_{\text{chunks}} \cdot \text{chk\_size})^{2} \Big)
\]

During decoding, only the current query vector is needed for attention, giving the following formula:

\[
\text{Mem}_{\text{dense\_decode}} \;=\;
2 \,\cdot \text{seq\_len} \,\cdot d_{\text{emb}}^{(KV)}
\;+\; \text{seq\_len}
\;+\; 2 \,\cdot d_{\text{emb}}^{(Q)}
\]
\[
\text{Mem}_{\text{APCE\_decode}} \;=\;
2 \, \cdot (\text{n}_{\text{chunks}} \cdot \text{chunk\_size}) \, \cdot d_{\text{emb}}^{(KV)}
\;+\; (\text{n}_{\text{chunks}} \cdot \text{chunk\_size})
\;+\; 2 \, d_{\text{emb}}^{(Q)}
\]
where the second term in both decoding formulas corresponds to the attention vector.

\subsection{Memory Footprint Analysis. }
We analyze the memory footprints for APCE and Full Dense for a single self-attention layer from the Llama-3.2-3B model having the embedding dimensions $d_{emb}^{(Q)}$=3072 for Q and output; $d_{emb}^{(KV)}$=1024 for the K and V vectors. We show the results across the 8k, 20k, and 30k context lengths when the number of selected chunks is equal to 70\% of the available chunks. Table \ref{tab:memory} shows that with 70\% of the available chunks selected at 800 token chunk size, APCE can improve the memory efficiencies of the KV cache and prefill attention by 32.8\% and 55.6\%, respectively. Here we make two important remarks: (1) Realizing the theoretical gains for attention operation would depend on how the attention kernel is implemented, where an ideal implementation would yield an attention map whose whose dimension equals the product of n$_{\text{chunks}}$  and chunk\_size; (2) During decoding, techniques like sliding window attention can also reduce the attention map size of the dense operation. APCE's merit remains that its input chunk selection facilitates controlling which KV vectors to use for self-attention.

\begin{table}[t!]
\centering
\caption{Theoretical Memory Analysis for self-attention using the Full-Dense and APCE in Llama-3.2-3B-Instruct \cite{dubey2024llama} on FP16 format. APCE assumes 70\% chunk selection at 800 token chunk size}
\vspace{2ex}
\label{tab:memory}
\begin{tabular}{c|c|ccc}
\toprule
{Context Length} & {Method} & KV-cache (MB) & Prefill Attn (MB) & Decode Attn (MB) \\
\midrule
\multirow{2}{*}{8k}  & Dense      &  32.40           &  260.80            &   32.43           \\
                     & APCE (70\%) &  21.88           &  147.31           &   21.90            \\
\midrule
\multirow{2}{*}{20k} & Dense      &  78.56           &  1060.0            &    78.61           \\
                     & APCE (70\%) &  56.25           &  620.51            &   56.29           \\
\midrule
\multirow{2}{*}{30k} & Dense      &  116.89          &  2120.0            &    116.96          \\
                     & APCE (70\%) &  75.00           &  1003.12            &      75.05        \\
\bottomrule
\end{tabular}
\end{table}





\section{Limitations }


We implemented a vanilla version of APCE for the purpose of validating our hypothesis regarding \textit{ContextRot} and how semantic-based input chunk selection can improve performance while enhancing memory efficiency. However, this vanilla implementation restricts understanding the full potential of APCE with regards to total generation time speedups, forcing a -- \textit{potentially unnecessary} -- trade-off between performance and efficiency. Though our approach remains independent from underlying programming environments like CUDA, having supporting, customized kernel implementations for APCE's self-attention can provide more clarity on APCE's total generation time performance. On the other hand from a performance evaluation standpoint, while the current study spans a range of context lengths, extending evaluation to the upper end of the long-context distribution could further reinforce the strength of the claimed arguments.

\end{document}